\documentclass[11pt,a4paper]{article}
\usepackage[hyperref]{eacl2021}
\usepackage{times}
\usepackage{latexsym}

\usepackage{graphicx}
\usepackage{amsmath}
\usepackage{float}
\usepackage{booktabs}
\usepackage{balance}
\usepackage{booktabs, multirow} % for borders and merged ranges
\usepackage{soul}% for underlines
\usepackage{color, colortbl}
\definecolor{LightCyan}{rgb}{0.88,1,1}
\usepackage{booktabs, multirow} % for borders and merged ranges
\usepackage{soul}% for underlines
\usepackage{changepage,threeparttable} % for wide tables

\usepackage{array}
\newcommand{\PreserveBackslash}[1]{\let\temp=\\#1\let\\=\temp}
\newcolumntype{C}[1]{>{\PreserveBackslash\centering}p{#1}}
\newcolumntype{R}[1]{>{\PreserveBackslash\raggedleft}p{#1}}
\newcolumntype{L}[1]{>{\PreserveBackslash\raggedright}p{#1}}

\newcommand{\method}{\textsc{ResPer}}

\usepackage{microtype}

\aclfinalcopy % Uncomment this line for the final submission
\def\aclpaperid{42} %  Enter the acl Paper ID here

\title{\method{} : Computationally Modelling Resisting Strategies in Persuasive Conversations.}

\author{Ritam Dutt$^{*,1}$, Sayan Sinha$^{*,2}$, Rishabh Joshi$^{1}$, Surya Shekhar Chakraborty$^{3}$,\\
\textbf{Meredith Riggs$^{1}$, Xinru Yan$^{1}$, Haogang Bao$^{1}$, Carolyn Penstein Ros\'{e}$^{1}$} \\
\\
$^{1}$Carnegie Mellon University, $^{2}$Indian Institute of Technology Kharagpur, $^{3}$Zendrive Inc\\
\texttt{\{rdutt,rjoshi2,mriggs,xinruyan,haogangb,cprose\}@cs.cmu.edu},\\
\texttt{sayan.sinha@iitkgp.ac.in, suryaschak@gmail.com}
}

\date{}

\begin{document}
\maketitle

\begin{abstract}
Modelling persuasion strategies as predictors of task outcome has several real-world applications and has received considerable attention from the computational linguistics community. However, previous research has failed to account for the resisting strategies employed by an individual to foil such persuasion attempts. Grounded in prior literature in cognitive and social psychology, we propose a generalised framework for identifying resisting strategies in persuasive conversations. We instantiate our framework on two distinct datasets comprising persuasion and negotiation conversations. We also leverage a hierarchical sequence-labelling neural architecture to infer the aforementioned resisting strategies automatically. Our experiments reveal the asymmetry of power roles in non-collaborative goal-directed conversations and the benefits accrued from incorporating resisting strategies on the final conversation outcome. We also investigate the role of different resisting strategies on the conversation outcome and glean insights that corroborate with past findings. We also make the code and the dataset of this work publicly available at \url{https://github.com/americast/resper}.

% for different scenarios and glean several insights. 

% Persuasion plays a prominent role in daily human interaction and has received 

\end{abstract}

\section{Introduction}

%\rd{First to propose a computational model of resisting strategies for persuasion.}
\let\thefootnote\relax\footnotetext{$^{*}$ denotes equal contribution}

Persuasion is pervasive in everyday human interactions. People are often exposed to scenarios that challenge their existing beliefs and opinions, such as medical advice, election campaigns, and advertisements \cite{knobloch2009looking, bartels2006priming, speck1997predictors}.
Of late, huge strides have been taken by the Computational Linguistics community to advance research in persuasion. Some seminal works include identifying persuasive strategies in text \cite{diyi-yang-persuasive} and conversations \cite{persuasion-for-good-2019}, investigating the interplay of language and prior beliefs on successful persuasion attempts \cite{durmus-cardie-2018-exploring, longpre-etal-2019-persuasion}, and generating persuasive dialogues \cite{munigala2018persuaide}.

However, a relatively unexplored domain by the community is the investigation of resisting strategies employed to foil persuasion attempts. As succinctly observed by Miller (1965): ``In our daily lives we are struck not by the ease of producing attitude change but by the rarity of it.''  Several works in cognitive and social psychology \cite{fransen2015strategies, zuwerink2003strategies} have put forward different resisting strategies and the motivations for the same. However, so far,  there has not been any attempt to operationalise these strategies from a computational standpoint. We attempt to bridge this gap in our work.  

We propose a generalised framework, grounded in cognitive psychology literature, for automatically identifying resisting strategies in persuasion oriented discussions.  We instantiate our framework on two publicly available datasets comprising persuasion and negotiation conversations to create an annotated corpus of resisting strategies. 

Furthermore, we design a hierarchical sequence modelling framework, that leverages the conversational context to identify resisting strategies automatically. Our model significantly outperforms several neural baselines, achieving a competitive macro-F1 score of 0.56 and 0.66 on the persuasion and negotiation dataset, respectively. 

We refer to our model as \method{}, which is not only an acronym for \textbf{Res}isting \textbf{Per}suasion, but also a play on the word ESPer: a person with extrasensory abilities. The name is apt since we observe that incorporating such resisting strategies could provide additional insight on the outcome of the conversation. In fact, our experiments reveal that the resisting strategies are better predictors of conversation success for the persuasion dataset than the strategies employed by the persuader. We also observe that the buyer's strategies are more influential in negotiating the final price. Our findings highlight the asymmetric nature of power roles arising in non-collaborative dialogue scenarios and form motivation for this work.

\section{Related Works}

The use of persuasion strategies to change a person's view or achieve a desired outcome finds several real-world applications, such as in election campaigns \cite{knobloch2009looking, bartels2006priming}, advertisements \cite{speck1997predictors}, and mediation \cite{cooley1993classical}. Consequently, several seminal NLP research have focused on operationalising and automatically identifying persuasion strategies \cite{persuasion-for-good-2019}, propaganda techniques \cite{propaganda-1}, and negotiation tactics \cite{yiheng_sigdial}, as well as the impact of such strategies on the outcome of a task  \cite{diyi-yang-persuasive, NegData, joshi2021dialograph}. However, there is still a dearth of research from a computational linguistic perspective investigating resisting strategies to foil persuasion. 

Resisting strategies have been widely discussed in literature from various aspects such as marketing \cite{heath2017beating}, cognitive psychology \cite{zuwerink2003strategies},  and political communication \cite{fransen2015typology} . Some notable works include the identification and motivation of commonly-used resisting strategies \cite{fransen2015strategies, zuwerink2003strategies}, the use of psychological metrics to predict resistance \cite{san2019role, ahluwalia2000examination}, and the design of a framework to measure the impact of resistance \cite{tormala2008new}. However, these works have mostly relied on qualitative methods, unlike ours, which adopts a data-driven approach.  We propose a generalised framework to characterise resisting strategies and employ state-of-the-art neural models to infer them automatically. Thus our work can be considered complementary to past research. 

The closest semblance to our work in NLP literature ties in with argumentation, be it essays \cite{argumentative-essays-2}, debates \cite{argumentative-debates}, or discussions on social media platforms \cite{argumentation-wikipedia, argumentative-cmv}. Such works have revolved mostly on analysing argumentative strategies and their effect on others.

Recently, \citet{resist-argumentative-persuasion} demonstrated that incorporating the personality traits of the resistor was influential in determining their resistance to persuasion. Such an observation acknowledges the power vested in an individual to resist change to their existing beliefs. Our work exhibits significant departure from this because we explicitly characterise the resisting strategies employed by the user.
% instead of leveraging post behaviour as a proxy for personality traits
Moreover, our work focuses on the general domain of non-collaborative task-oriented dialogues, where several non-factual resisting strategies are observed, making it distinctly different from argumentation \cite{galitsky2018argumentation}. We assert that focusing on both parties is imperative to get a complete picture of persuasive conversations. 

\begin{table*}[t]
\small
\centering

\begin{tabular}{p{0.16\textwidth} p{0.37\textwidth}p{0.38\textwidth}}
\toprule
Resisting Strategy & Persuasion (P4G) & Negotiation (CB)\\ \midrule

Source Derogation & Attacks/doubts the organisation's credibility. & Attacks the other party or questions the item.\\ 

& \textit{My money probably won't go to the right place}& \textit{Was it new denim, or were they someone's funky old worn out jeans?} \\ \hline

Counter Argument & Argues that the responsibility of donation is not on them or refutes a previous statement.  & Provides a non-personal argument/factual response to refute a previous claim or to justify a new claim. \\ 

&\textit{There are other people who are richer} & \textit{It may be old, but it runs great. Has lower mileage and a clean title.}\\ \hline

Personal Choice & Attempts to saves face by asserting  their personal preference such as their choice of charity and their choice of donation. & Provides a personal reason for disagreeing with the current situation or chooses to agree with the situation provided some specific condition is met.\\
& \textit{I prefer to volunteer my time} & \textit{I will take it for \$300 if you throw in that printer too.}\\  \hline

Information Inquiry & Ask for factual information about the organisation for clarification or as an attempt to stall. & Requests for clarification or asks additional information about the item or situation. \\
& \textit{What percentage of the money goes to the children?} & \textit{Can you still fit it in your pocket with the case on?}\\ \hline

Self Pity & Provides a self-centred reason for not being able/willing to donate at the moment. & Provides a  reason (meant to elicit sympathy) for disagreeing with the current terms.\\
& \textit{I have my own children} & \textit{\$130 please I only have \$130 in my budget this month.}\\ \hline

Hesitance & Attempts to stall the conversation by either stating they would donate later or is currently unsure about donating.  & Stalls for time and is hesitant to commit; specifically, they seek to further the conversation and provide a chance for the other party to make a better offer. \\
&\textit{Yes, I might have to wait until my check arrives.} & 
\textit{Ok, would you be willing to take \$50 for it?}\\ \hline

Self-assertion & Explicitly refuses to donate without even providing a factual/personal reason & Asserts a new claim or refutes a previous claim with an air of finality/ confidence.\\ &\textit{Not today} & \textit{That is way too little.}\\ \bottomrule
\end{tabular}
\caption{Framework describing the resisting strategies for persuasion (P4G) and negotiation (CB) datasets. We emphasise that Information Inquiry is not a resisting strategy for CB.  Examples of each strategy are italicised.}

\label{tab:framework}
\end{table*}

\section{Framework}
% \rd{Type of strategies/Dataset: Depends on the dataset, hence we can restrict ourselves to the properties which are present in the dataset. }

In this section, we describe the datasets, the resisting strategies employed, and the annotation framework to instantiate the strategies. 

% we present our annotation framework for identifying resisting strategies in persuasive and negotiation discussions and discuss the datasets employed.

\subsection{Dataset Employed}
 We choose persuasion-oriented conversations, rather than essays or advertisements \cite{diyi-yang-persuasive}, since we can observe how the participants respond to the persuasion attempts in real-time. To that end, we leverage two publicly available corpora on persuasion \cite{persuasion-for-good-2019} and negotiation \cite{NegData}. We refer to these datasets as ``Persuasion4Good'' or P4G and ``Craigslist Bargain'' or CB hereafter.

\textbf{P4G} comprises conversational exchanges between two anonymous Amazon Mechanical Turk workers with designated roles of the persuader, ER and persuadee, EE.  ER  had to convince EE \space to donate a part of their task earnings to the charity \emph{Save the Children}. We investigate the resisting strategies employed only by EE \space in response to the donation efforts. We emphasise that the conversational exchanges are not scripted, and the task is set up so that a part of EE's earnings is deducted if they agree to donate. Since there is a monetary loss at stake for EE, we expect them to resist.

\textbf{CB} consists of simulated conversations between a buyer (BU) and a seller (SE) over an online exchange platform. Both are given their respective target prices and employ resisting strategies to negotiate the offer.

We choose these datasets since they involve non-collaborative goal-oriented dialogues. As a result, we can definitively assess the impact of different resisting strategies on the goal.

\begin{table}[!htp]\centering
\caption{Description  for the Persuasion (P4G) \cite{persuasion-for-good-2019} and Negotiation (CB) \cite{NegData} datasets}\label{tab:dataset properties}
\resizebox{\linewidth}{!}{\begin{tabular}{lrrr}\toprule
\textbf{Properties} &\textbf{P4G} &\textbf{CB} \\\midrule
\# of conversations &530 & 800 \\
Max \# of utterances/conversation &76 & 44 \\
Avg \# of utterances/conversation &36.34 & 11.94 \\
Max \# of tokens/utterance &90 &93 \\
Avg \# of tokens/utterance &11.03 &14.62 \\
Vocabulary size &6137 & 5370 \\
\bottomrule
\end{tabular}}
\end{table}

\subsection{Framework Description}
In this subsection, we briefly describe the resisting strategies commonly referenced in social and cognitive psychology literature. This enables us to design a unified framework for the two datasets, built upon common underlying semantic themes. 
% We ground our framework for identifying resisting strategies in persuasive conversations in pre-existing literature of cognitive psychology and social psychology \cite{fransen2015strategies,zuwerink2003strategies}.
\citet{fransen2015strategies} identified 4 major clusters 
of resisting strategies, namely \textbf{contesting} \cite{wright1975factors, zuwerink2003strategies, abelson1967negative}, \textbf{empowerment} \cite{zuwerink2003strategies, sherman1980attitude}, \textbf{biased processing} \cite{ahluwalia2000examination}, and \textbf{avoidance} \cite{speck1997predictors}. Each individual category can be subdivided into finer categories showcased in italics henceforth. 
%These 4 clusters can be further subdivided into more fine-grained strategies. 

Contesting refers to attacking either the source of the message (\textit{Source Derogation}) or its content (\textit{Counter Argumentation}).  A milder form of contesting involves seeking clarification or information termed \textit{Information Inquiry}. Prior work has shown a positive association between working knowledge and one's ability to resist persuasion \cite{wood1988communicator, luttrell2020attitude}. Therefore, \textit{Information Inquiry} can be interpreted as a form of resistance where the resistor seeks to satisfy their doubts because they are sceptical of the persuader's intents or messages. This is prominent in certain conversations in P4G where a sceptical EE questions the charity's legitimacy.

% \rd{Does not follow: Previous work \cite{zuwerink2003strategies} has focused on observing resisting strategies in argumentative essays or advertisements, which lack a conversational back-and-forth; hence we posit \textit{Information Inquiry} as a form of resistance.}

Empowerment strategies encompass %several multi-faceted attempts to direct the flow of the conversation towards oneself. These 
reinforcing one's personal preference to refute a claim (\textit{Attitude Bolstering}) \cite{sherman1980attitude}, attempting to arouse guilt in the opposing party (\textit{Self Pity}) \cite{self-pity2, self-pity1}, stating one's wants outright (\textit{Self Assertion}) \cite{zuwerink2003strategies}, or seeking validation from like-minded people (\textit{Social Validation}) \cite{fransen2015strategies}. Overall, empowerment strategies drive the discussion towards the resistor's self as opposed to attacking the persuader. 

Biased processing mitigates external persuasion by selectively processing information that conforms with one's opinion or beliefs %(\textit{Reducing Impact} and \textit{Weighing Attributes})
\cite{fransen2015strategies}. For simplicity, we subsume strategies that denote personal preference, namely \textit{Attitude Bolstering} and \textit{Biased Processing}, into a unified category \textit{Personal Choice}. We refrain from incorporating \textit{Self Assertion} into the \textit{Personal Choice} category since it deals with bolstering one's confidence and not one's opinions or attitudes. The subtle difference is highlighted in Table \ref{tab:framework}. 

Avoidance strategies distance the resistor from persuasion, either physically or mechanically, or refuse to engage in topics that induce cognitive dissonance \cite{fransen2015strategies}. However, in the context of task-oriented conversations, wherein participants are expected to further a goal, avoidance often manifests as \textit{Hesitance} to commit to the current situation.

We identify seven major resisting strategies across the datasets, namely \textit{Source Derogation, Counter Argumentation, Information Inquiry, Personal Choice, Self Pity, Hesitance,} and \textit{Self Assertion}. Since the datasets comprise two-party conversations between strangers, \textit{Social Validation}, which requires garnering the support of others, was absent. We now describe how these resisting strategies were instantiated in the following section.

%Please add the following packages if necessary:
%\usepackage{booktabs, multirow} % for borders and merged ranges
%\usepackage{soul}% for underlines
%\usepackage[table]{xcolor} % for cell colors
%\usepackage{changepage,threeparttable} % for wide tables
%If the table is too wide, replace \begin{table}[!htp]...\end{table} with
%\begin{adjustwidth}{-2.5 cm}{-2.5 cm}\centering\begin{threeparttable}[!htb]...\end{threeparttable}\end{adjustwidth}
\subsection{Instantiating the Resistance Framework}

We emphasise that although the description and meaning of a strategy remain the same across the two datasets, their semantic interpretation depends on the context. For example, scepticism towards the charity in P4G and criticism of the product in CB are instances of \textit{Source Derogation}. This is because ER represents the charity, whereas the seller is being accused of selling an inferior product. Likewise, we instantiate the predicates for the remaining six resisting strategies for the two datasets, with examples in Table \ref{tab:framework}. 

We label the utterances of persuadee (EE) in P4G and the buyers (BU) and sellers (SE) in CB with at least one of the seven corresponding resisting strategies, or `Not-A-Strategy' if none applies. The `Not-A-Strategy' label includes greetings, off-task discussions, agreement, compliments, or other tokens of approval. We acknowledge that an utterance can have more than one resisting strategy embedded in it. 
For example, the utterance ``The price is slightly high for used couches, would you come down to $240$ if I also picked them up?", is an instance of both \textit{ Personal Choice} and	\textit{Counter Argumentation}.

% For instance ``I know \textit{$<$brand$>$} does not have very high-quality items. What are the dimensions?" is an instance of both \textit{Source Derogation} and \textit{Information Inquiry}.
We also note that \textbf{\textit{Information-Inquiry} is not a resisting strategy for CB} since asking additional information/clarification is an expected behaviour before finalising a deal. We keep the label nevertheless to show comparison with P4G. We present the flowchart detailing the annotation framework in Figure \ref{fig:flowchart} of Appendix. 

\subsection{Annotation Procedure and Validation}

We describe the annotation procedure for both the CB and P4G dataset here and its subsequent validation. For CB, three authors independently annotated five random conversations adhering to the flowchart. If the conversations chosen were simple or had few labels, a new set of 5 conversations were taken up. This constitutes one round. After each round, the Fleiss Kappa score was computed, and the authors discussed to resolve the disagreements and revise the flowchart. Then began the next round on a new set of 5 random conversations. For CB, 5 rounds of revision were carried out over 24 conversations, until a high Fleiss kappa (0.790) \cite{fleiss1971measuring}  was obtained. Finally, the three authors independently went ahead and annotated approximately 250 distinct conversations, yielding a corpus of 800 CB conversations. Our annotation procedure requires a rigorous reliable refinement phase but a comparatively faster annotation phase by dividing the annotation between the authors. Thus the conversations annotated by each author were mutually exclusive. Similarly, for P4G dataset, four authors annotated 3 conversations per round, since a conversation in P4G was comparatively longer. 4 rounds of revision across 12 conversations was done to achieve the final kappa-score of 0.787. The four authors then went ahead and divided the task of annotating the 500 conversations amongst themselves.  %The first authors in the paper was involved in designing the annotation framework and the annotation for both CB and P4G. They were assisted by five others,  3 for  P4G and 2 for CB. 
We show an annotated conversation snippet for the two datasets in Table \ref{tab:annotation-snippet}.

% \textbf{Validating the annotation scheme:} Four authors of the paper annotated $530$ P4G conversations, whereas another three annotated $800$ CB conversations. The annotation scheme went through several revisions, eventually yielding a high Fleiss Kappa \cite{fleiss1971measuring} score of 0.787 and 0.790 for P4G and CB  respectively. The revised scheme was then used to annotate the remaining conversations. 
% We show an annotated conversation snippet in \ref{tab: example} of Appendix \ref{sec:appendix}.
\begin{table}[!t]
% \normalsize
\footnotesize
% \large
\centering
%  \scriptsize
\caption{Examples of annotation snippets for the Persuasion (P4G) and Negotiation (CB). The utterances of the EE and the SE are highlighted in cyan. Some strategies are shortened, like Info Inquiry, and Per Choice for Information Inquiry and Personal Choice.}
\label{tab:annotation-snippet}
\resizebox{0.47\textwidth}{!}{
\begin{tabular}{{p{0.02\textwidth}p{0.36\textwidth}p{0.1\textwidth}}}\toprule
\textbf{Role} &\textbf{Text} &\textbf{Strategy} \\\midrule
\multicolumn{3}{c}{\textbf{Negotiation (CB)}} \\
\midrule
\rowcolor{LightCyan}
SE& I have a wonderful phone for you if you are interested.& No Strategy \\
BU& I am interested. Did you just buy it?& Info inquiry \\
\rowcolor{LightCyan}
SE& I bought it two weeks ago but it just wasn't what I needed anymore.& No Strategy \\
BU& Would you be willing to work with the price?& Hesitance \\
\rowcolor{LightCyan}SE& Yes we can negotiate.& No Strategy \\
BU& If I come today would you accept \$56 I can bring it now?& Per Choice \\
\rowcolor{LightCyan}SE& How about 65 and I can deliver it to you now?& Per Choice \\
BU& Can you go \$60 Kind of all I have right now ?& Self Pity \\
\rowcolor{LightCyan}SE& Yes I can.& No Strategy \\
% 0 &Does this TV really have a skewed wooden frame on it? &Source Derogation \\
% 1 &Hi there no there is no wooden frame &Not-A-Strategy \\
% 0 &okay because in the picture it looks cattywompous. what do you say to \$350? &Hesitance \\
% 1 &the tv is in perfect condition the photo is just slightly askew as it is in perfect condition i really do not want to go that low, &Counter Argumentation \\
% 0 &does the doll house come with it? &Personal Choice \\
% 1 &no i'm afraid not. but i can say there is absolutely nothing wrong with it, as if it just came out of the box today. &Not-A-Strategy \\
% 0 &what do you say to \$360 then? &Hesitance \\
% 1 &the tv retails for over 600 dollars i really don't want to go any lower then \$375 &Hesitance \\
% 0 &hmmmm. you drive a hard bargain. okay \$375 and the dollhouse, please. i have to have the doll house. i want the cattywompous setup! &Personal Choice \\
% 1 &HaHa sorry the dollhouse would be an extra \$100 dollars &Self Assertion \\
% 0 &nooooooo. you're breaking my heart mister. okay \$400 and the dollhouse comes with? make my cattywompous life complete please sir. &Personal Choice \\
% 1 &sorry i would have to get my daughter a new one therefore i can not negotiate on the dolhouse &Self Pity \\
% 0 &sigh. okay fine. \$375 and now mr whiskers is homeless. i hope you're happy mister. &Not-A-Strategy \\
\midrule
\multicolumn{3}{c}{\textbf{Persuasion (P4G)}} \\
\midrule 
ER &Hello, Save the Children looks like an interesting organisation. &- \\
\rowcolor{LightCyan}EE &i would like to know more about it &Info Inquiry \\
% 0 &Save the Children is an international non-governmental organisation that helps support children in developing countries. &- \\
% 1 &and do they have a website ? &Information-inquiry \\
% 0 &They do, it is URL &- \\
.. & .. & ..\\
\rowcolor{LightCyan}EE &thanks i will definitely check it out &Hesitance \\
ER &They also promote children's rights and provide relief when needed. &- \\
\rowcolor{LightCyan}EE &and where does the money go if i do donate ? &Info Inquiry \\
\rowcolor{LightCyan}EE &Straight to the organisation? &Info Inquiry \\
ER &Yes, it goes straight to the organisation, where it can be used to help many children. &- \\
\rowcolor{LightCyan}EE &because some organisations do not divide the money properly &Source Derogation \\
ER &This organisation has been checked by some groups, and they divide the money properly. &- \\
.. & .. & ..\\
% ER &There's nothing bad about donating to this organisation. &- \\
\rowcolor{LightCyan}EE &I will certainly consider it &No Strategy \\
% 0 &I have read about children dying from hunger, and the donations made to this group have saved many others like them. &- \\
% 1 &that is just wonderful, i love hearing stories like this &Not-A-Strategy \\
% 0 &Yeah me too, children in Syria have to grow up facing threat of violence everyday, in the first two months of 2018 there were 1,000 children who were killed or injured. &- \\
% 1 &Oh man i was not aware of this &Not-A-Strategy \\
% 0 &Yeah it's pretty bad, and the donations can make a big difference, just think about all the snack food you buy, you could skip a week or two and make a donation from that. &- \\
% 1 &i hardly eat snacks but you have a point &counter-argumentation \\
% 0 &I hope you make a really good donation as the donations will go to a trusted fund where it will help many children. &- \\
% EE &i sure will &Not-A-Strategy \\
\bottomrule
\end{tabular}}
\end{table}

\subsection{Dataset Statistics}

The P4G and CB datasets comprise $530$ and $800$ labelled conversations, respectively, spanning an average of 37 and 12 utterances per conversation. 

The datasets cover two distinct persuasion scenarios and also illustrate the rights and obligations shown by the participants. For example, in P4G, EE comes into the interaction blind and is unaware of the donation attempt. We encounter several conversations where EE is willing to donate since it resonates with their beliefs, and no resisting strategies are observed. However, for CB, the participants received prior instructions to negotiate a deal, and hence resisting strategies were more prominent. We present the frequency distribution of the seven strategies in Table \ref{tab:dataset_freq}. We observe that the distributions of strategies are skewed for both the datasets and is more pronounced for P4G, where `Not-A-Strategy' accounts for the lion's share. We also see that the buyer exhibits more resisting strategies than the seller highlighting the asymmetric role of the two participants. 

% We also observe that the vocabulary of the two datasets is widely different. The shared vocabulary constitutes only 37\% and 43\% of the individual vocabulary for P4G and CB, highlighting a prominent lexical shift between the two domains. 

Nevertheless, we reiterate that the resisting strategies we propose are applicable for both the domains. In the next section, we propose the framework to infer such strategies automatically. 

% We notice that \textit{Information Inquiry}  comprises the highest number of utterances, followed by \textit{Personal Choice}. However, it is to be noted that \textit{Information Inquiry}, when grounded to the Negotiation Dataset \cite{NegData}, is not considered a resisting strategy. It highlights the willingness to learn more about the item being sold or purchased, unlike the persuasion dataset where it symbolises casting doubts. %Hence, we can see, it features a lower variance among strategies.
% % \SayanStart Some explanation \SayanEnd.
% Table  \ref{tab: data properties} highlights the fact that the Negotiation dataset is around 2.5 times larger than the Persuasion dataset, explaining the reason behind certain results as obtained in the subsequent sections. 

%Please add the following packages if necessary:
%\usepackage{booktabs, multirow} % for borders and merged ranges
%\usepackage{soul}% for underlines
%\usepackage[table]{xcolor} % for cell colors
%\usepackage{changepage,threeparttable} % for wide tables
%If the table is too wide, replace \begin{table}[!htp]...\end{table} with
%\begin{adjustwidth}{-2.5 cm}{-2.5 cm}\centering\begin{threeparttable}[!htb]...\end{threeparttable}\end{adjustwidth}
\begin{table}[h]\centering
\caption{Proportion of resisting strategies (in \%)for the
Persuasion (P4G) and Negotiation (CB) dataset. The strategies are observed only for the persuadee (EE) in P4G and for both buyer (BU)
and seller (SE) in CB.}\label{tab:dataset_freq}
\footnotesize
\resizebox{0.47\textwidth}{!}{
\begin{tabular}{lrrrr}\toprule
\multirow{2}{*}{\textbf{Strategy}} %&\multirow{2}{*}{\textbf{Persuasion}} 
&\multicolumn{1}{c}{\textbf{Persuasion (P4G)}}
&\multicolumn{2}{c}{\textbf{Negotiation (CB)}} \\\cmidrule{2-4}
& \textbf{EE}&\textbf{BU} &\textbf{SE} \\\midrule
Source Derogation &2.16 &7.61 &0.44 \\
Counter Argument &2.28 &3.74 &6.06 \\
Personal Choice &2.52 &9.43 &8.49 \\
Information Inquiry &7.19 &18.27 &0.38 \\
Self Pity &1.58 &4.66 &0.34 \\
Hesitance &1.76 &15.78 &9.14 \\
Self-assertion &0.94 &2.20 &5.05 \\
Not a strategy &81.56 &38.30 &70.09 \\
\bottomrule
\end{tabular}}
% \vspace{-0.7cm}
\end{table}

%Please add the following packages if necessary:
%\usepackage{booktabs, multirow} % for borders and merged ranges
%\usepackage{soul}% for underlines
%\usepackage[table]{xcolor} % for cell colors
%\usepackage{changepage,threeparttable} % for wide tables
%If the table is too wide, replace \begin{table}[!htp]...\end{table} with
%\begin{adjustwidth}{-2.5 cm}{-2.5 cm}\centering\begin{threeparttable}[!htb]...\end{threeparttable}\end{adjustwidth}

\begin{figure*}
    \centering
    \includegraphics[scale=0.32]{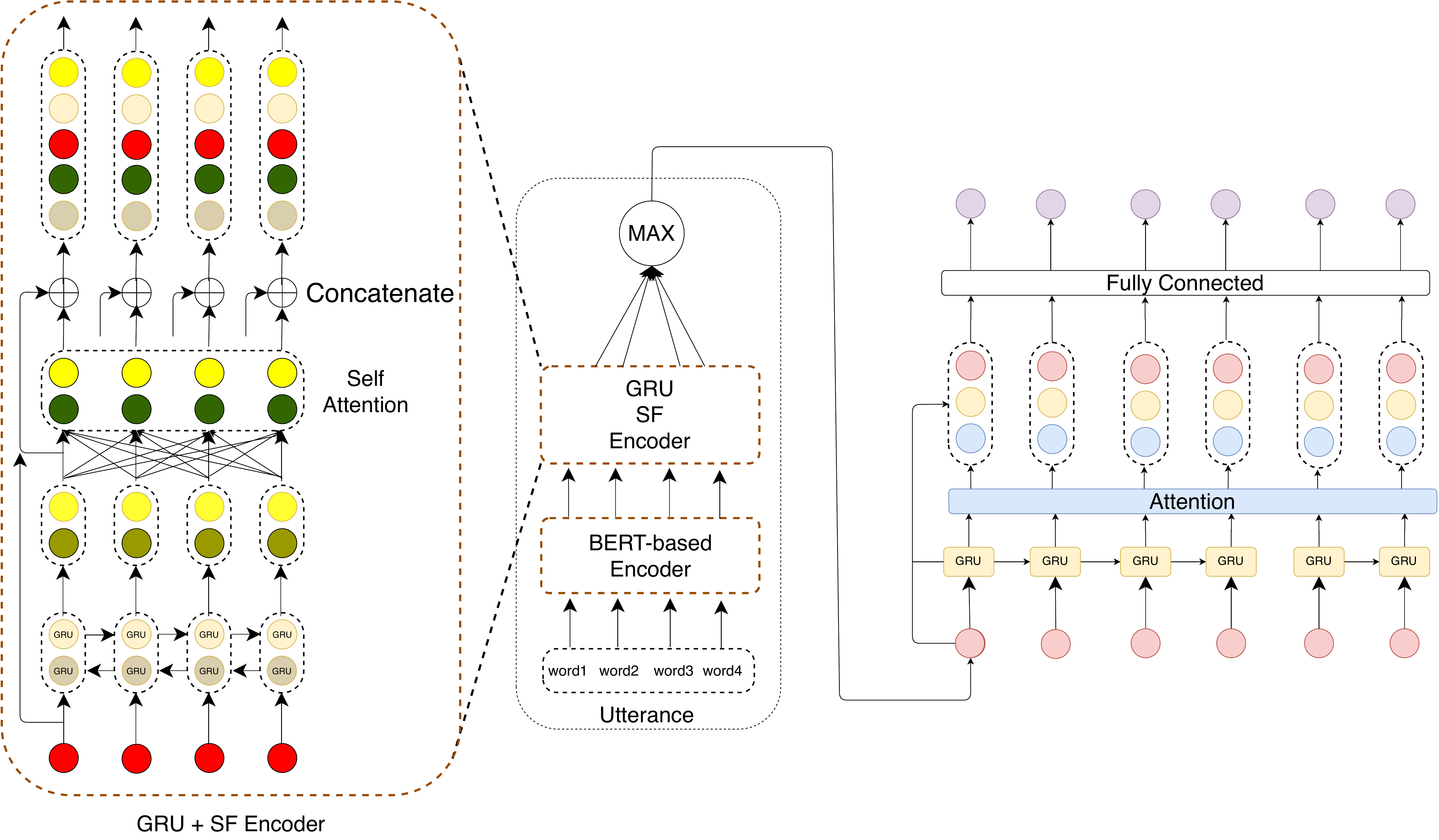}
    \caption{A diagram illustrating how \method{} works. The encoder shown on the left takes the BERT representations of a token as input and passes it through a BiGRU layer followed by Self Attention. The outputs from BERT, BiGRU and self-attention are then concatenated to form the output. Max-pooling over this output yields the corresponding utterance embedding. This utterance representation is passed through a uni-directional GRU followed by Masked-Self-Attention and fusion to yield the contextualised utterance embedding.}
    \label{fig:model}
\end{figure*}

\section{Methodology}
\label{sec:models}
In this section, we describe the methodology adopted for inferring the resisting strategies in persuasion dialogues and how they can be leveraged to determine the dialogue's outcome. 

\subsection{Resisting Strategy prediction}

We model the task of identifying resisting strategies as a sequence labelling task. We assign each utterance in the dialogues with a label representing either one of the seven resisting strategies or \emph{Not-A-Strategy}. \footnote{We acknowledge that an utterance can have multiple labels. However, such utterances comprise only 1.2\% and 3.85\% of the P4G and the CB datasets, respectively. In such cases, the label is randomly selected.}

% Given a dialogue with $n$ utterances, $D = [u_{1},u_{2}, ..., u_{n}]$, we assign labels $y_{1}, y_{2} ... y_{n}$ where $y_{i} \in Y$ represents one of 7 possible resisting strategies or `Not-A-Strategy'. 

% Although we acknowledge that an utterance can have multiple resistance labels, they comprise only \rd{1.2\%, verify this} of for Persuasion and 3.85\% for Negotiation respectively. We thus design our model to predict a single label for each utterance, wherein the selected label is chosen at random. 

Since the resisting strategies, by definition, occur in response to the persuasion attempts, our model architecture needs to be cognizant of the conversational history. To that end, we adopt a hierarchical neural network architecture, similar to \citet{HiGRU}, to infer the corresponding resisting strategy. The architecture leverages the previous conversational context in addition to the current contextualised utterance embedding. Our choice is motivated by the recent successes of hierarchical sequence labelling frameworks in achieving state-of-the-art performance on several dialogue-oriented tasks. Some myriad examples include emotion recognition \cite{majumder2019dialoguernn, HiGRU}, dialogue act classification \cite{chen2018DAP,raheja-tetreault-2019-dialogue}, face act prediction \cite{dutt2020keeping}, open domain chit-chat \cite{persona_chat, amused} and the like.
We hereby adopt this as the foundation architecture for our work and refer to our instantiation of the architecture as \method{}.

\noindent{\textbf{Architecture of \method{}:}} An utterance $u_{j}$ is composed of tokens $[w_0, w_1, ..., w_K]$ represented by their corresponding embeddings $[e(w_0), e(w_1),..., e(w_K)]$. In \method{}, we obtain these using a pre-trained BERT model \cite{devlin2019bert}.
We pass these contextualised word representations through a bidirectional GRU to obtain the forward $\overrightarrow{h_{k}}$ and backward $\overleftarrow{h_{k}}$ hidden states of each word, before passing them into a Self-Attention layer. This gives us the corresponding attention outputs, $\overrightarrow{ah_{k}}$ and $\overleftarrow{ah_{k}}$ as described below.
\begin{equation*}
\begin{split}
      \overrightarrow{h_{k}} = \operatorname{GRU}\left(e\left(w_{k}\right), \overrightarrow{h_{k-1}}\right) \\
      \overleftarrow{h_{k}}= \operatorname{GRU}\left(e\left(w_{k}\right), \overleftarrow{h_{k+1}}\right)\\
    \overrightarrow{ah_{k}} = SelfAttention(\overrightarrow{h_{k}})\\
    \overleftarrow{ah_{k}} = SelfAttention(\overleftarrow{h_{k}})
\end{split}
\label{eq:SA}
\end{equation*}
Finally, we concatenate the contextualised word embedding with the GRU hidden states and Attention outputs in the \textit{fusion layer} to obtain the final representation of the word $e_c(w_k)$. We represent the bias as $b_w$. Here, We perform max-pooling over the fused word embeddings to obtain  the $j^{th}$ utterance embedding, $e(u_j)$. 
\begin{equation*}
    \begin{split}
      e_c(w_k) = \operatorname{tanh}(&W_w[\overrightarrow{ah_{k}}; \overrightarrow{h_{k}}; e(w_k); \overleftarrow{h_{k}}; \overleftarrow{ah_{k}}] + b_w)\\
      e(u_j) =& \operatorname{max}(e_c(w_1), e_c(w_2),... e_c(w_K))
    \end{split}
\end{equation*}

% Formally, 
% \begin{equation*}
%     \begin{split}
%       \overrightarrow{h_{k}} =& \operatorname{GRU}\left(e\left(w_{k}\right), \overrightarrow{h_{k-1}}\right) \\
%       \overleftarrow{h_{k}} =&\operatorname{GRU}\left(e\left(w_{k}\right), \overleftarrow{h_{k+1}}\right)\\
%       \overrightarrow{ah_{k}} =&\operatorname{SelfAttention}(\overrightarrow{h_{k}})\\
%       \overleftarrow{ah_{k}} =& \operatorname{SelfAttention}(\overleftarrow{h_{k}})\\
%       e_c(w_k) = \operatorname{tanh}(&W_w[\overrightarrow{ah_{k}}; \overrightarrow{h_{k}}; e(w_k); \overleftarrow{h_{k}}; \overleftarrow{ah_{k}}] + b_w)\\
%       e(u_j) =& \operatorname{max}(e_c(w_1), e_c(w_2),... e_c(w_K))
%     \end{split}
% \end{equation*}

% \begin{equation}
% \label{eqn:one}

% \end{equation}
We use a unidirectional GRU and Masked Self-Attention to encode conversational context, to ensure that the prediction for the $j^{th}$ utterance is not influenced by future utterances. Similarly, we calculate the contextualized representation of an utterance $e_c(u_j)$ using the conversation context.  We pass $e(u_{j})$ through a uni-directional GRU that yields 
the forward hidden state $\overrightarrow{H_{j}}$. Masked Self-Attention over the previous hidden states, yields $\overrightarrow{AH_{j}}$. We fuse $e(u_j)$, $\overrightarrow{H_{j}}$ and  $\overrightarrow{AH_{j}}$ before passing it through a linear layer with tanh activation to obtain  $e_c(u_j)$. %This ensures that current utterance is not influenced by future utterances. 
% \begin{equation*}
%     \begin{split}
%         \overrightarrow{H_{j}} =& \operatorname{GRU}\left(e\left(u_{j}\right), \overrightarrow{H_{j-1}}\right)\\
%         \overrightarrow{AH_{j}} =& \operatorname{MaskSelfAttention}(\overrightarrow{H_{j}})    
%     \end{split}
% \end{equation*}

% \begin{equation}
%     \label{eqn:two}
%     e_c(u_j) = \operatorname{tanh}(W_u[\overrightarrow{AH_{j}}; \overrightarrow{H_{j}}; e(u_j)] + b_u)
% \end{equation}

We project the final contextualised utterance embedding $e_c(u_j)$ onto the state space of resisting strategies. We apply softmax to obtain a probability distribution over the strategies, with  Negative Log-Likelihood (NLL) as the loss function to obtain the strategy loss.
%Given the true labels $y$ and the predicted labels $y'$, the loss is computed for all $n$ utterances in a conversation as:

% \begin{equation}
%     L_{f} = -\sum_{i=1}^{n} \sum_{y_j \epsilon Y}y_jlog(y'_{j})
% \end{equation}

% \SayanStart
% Shift this:
% \noindent\textbf{Architecture of Masked-\method{}:} In order to capture information from previous resisting strategies predicted by the model for utterances in a particular conversation, we add skip connections from the final softmax layer (where $y'$ is predicted) into the \textit{fusion layer}. The presence of such skip connections is determined by a binary \textit{Mask} vector, which represents the speaker of an utterance. The \textit{Mask} vector is a representative of the speaker in a conversation. Skip connections are only established when an utterance has value 1 in the \textit{Mask} vector. This ensures connections only across utterances from the same speaker. This helps skip the predicted vectors from conversations which naturally do not possess any resisting strategy.
% \SayanEnd

\subsection{Conversation Outcome prediction}
We further investigate the impact of resisting strategies on the outcome of the conversation. We represent a strategy as a fixed dimensional embedding initialised at random. We subsequently encode a sequence of strategies by passing them through a uni-directional GRU to obtain a final representation for the sequence. We project the representation onto a binary vector which encodes for the conversation outcome.  We apply softmax with NLL across all the conversations to obtain the outcome prediction loss.

%of resisting strategies Given a sequence of strategies, $[s_0, s_1, ..., s_N]$ and their corresponding embedding $[e(s_0),e(s_1), ..., e(s_N) ]$, for the $i^{th}$ conversation, we encode the sequence by passing it 
% % through a uni-directional GRU to obtain a final strategy representation ($e_s(c_i)$). Likewise, we obtain an encoding of the conversation text ($e_t(c_i)$) by passing the contextualized utterance embedding $e_c(u_j)$ through another uni-directional GRU and concatenate it with the strategy representation. We feed the concatenated representation through FC-layers with dropout and binarize the vector. We apply softmax with NLL across all the conversations to obtain the outcome prediction loss $L_{p}$. 
% \begin{equation*}
%     \begin{split}
%          e_s(c_i) =& \operatorname{GRU-Encode}([e(s_0), e(s_1), ..., e(s_N)])\\
%          e_t(c_i) =& \operatorname{GRU-Encode}([e_c(u_0), e_c(u_1), ..., e_c(u_N)])\\
%          st(i)= &\operatorname{tanh}(W_s[e_s(c_i)] + b_s)\\
%          text(i)= &\operatorname{tanh}(W_t[e_t(c_i)] + b_t)\\
%          o'_{i} =&\operatorname{softmax}(W_c[st(i):text(i)]+ b_c)\\
%          {L}_{p} = &-\sum_{i=1}^{n}o_ilog(o'_{i})\\
%     \end{split}
% \end{equation*}

\section{Experiments}
In this section, we describe the baselines and evaluation metrics. We present the experimental details of our model in Table \ref{tab:hyperparameters}.

\subsection{Baselines}

\noindent\textbf{Resisting strategy prediction:}
We experiment with standard neural baselines for text classification, which have also been used in classifying persuasion strategies, namely CNN \cite{kim2014convolutional, persuasion-for-good-2019} and BiGRU \cite{diyi-yang-persuasive}.  To ensure a fair comparison, we introduce pre-trained BERT-embeddings \cite{devlin2019bert} as input to the baselines, henceforth denoted as BERT-CNN and BERT-BiGRU. Furthermore, to inspect the impact of conversational history, we remove the conversational GRU from \method{} such that the utterance embedding $e(u_j)$ is directly used for prediction. We refer to this architecture as BERT-BiGRU-sf, since it employs self-attention(s) and fusion (f) on top of BERT-BiGRU. Finally, we experiment with the best performing HiGRU-sf model of \citet{HiGRU} as another baseline.

\noindent \textbf{Conversation success prediction:} 
The notion of conversation success depends on the choice of dataset. For P4G, we consider the resisting strategies to be successful if the persuadee (EE) refused to donate to charity.  For CB, we adopt the same notion of success as \citet{yiheng_sigdial}, namely when the seller (SE) can sell at a price greater than the median sale-to-list ratio $r$.  
\begin{equation}
\label{eqn:ratio}
r = \frac{\text{sale price} - \text{buyer target price}}{\text{listed price} - \text{buyer target price}}
\end{equation}

To observe the effect of conversation success, we experiment with
strategies of both the parties involved. For P4G, we encode separately (i) the persuasion strategies of ER as identified by \citet{persuasion-for-good-2019}, (ii) the resisting strategies employed by EE and (iii) both the persuasion and resisting strategies. Likewise, for CB, we encode the resisting strategies of only (i) the buyer (BU)  (ii) the seller (SE) (iii) both. These experiments would enable us to investigate which party has a greater influence on conversation success. 

\begin{table}[!htp]\centering
\caption{Here we describe the search-space of all the
hyper-parameters used in our experiments and describe
the search space we used to find the hyper-parameters.
$d_{h1}, d_{h2}$ represents the hidden dimensions of the Utterance GRU and the Conversation GRU.}\label{tab:hyperparameters}
\small
\begin{tabular}{lrrr}\toprule
\textbf{Hyper-parameter} &\textbf{Search space} &\textbf{Final Value} \\\midrule
learning-rate (lr) &1e-3 to 1e-5 &1e-4 \\
Batch-size &- &1 conversation \\
$\#$Epochs & $<100$ &30.8, 22 \\
lr-decay &- &0.5 every 20 epochs \\
$d_{h1}$ &- &1024 \\
$d_{h2}$ &- &300 \\
\bottomrule
\end{tabular}
\end{table}

% (i) only strategy embedding ($st(i)$) (ii) only the text embedding ($text(i)$)  (iii) concatenation of both strategy and text embeddings. We experiment with different architectures such as GRU, LSTM, and transformers to encode the particular sequences of strategies, and different pooling options like max-pooling, mean-pooling, or taking the last embedding of the sequence. We perform these experiments by encoding varying lengths of the conversation, namely 0.25, 0.5, 0.75 and 1.0.

\subsection{Evaluation metrics}
We adopt the same evaluation procedure for both the resisting strategy and the conversation outcome prediction task across the datasets. In either case, we perform five-fold cross-validation due to paucity of annotated data. We report performance in terms of the weighted\footnote{Weighted F1 Scores are calculated by taking the average of the F1 scores for each label weighted by the number of true instances for each label.} and macro F1-scores across the five folds. Our choice of the metric is motivated by the high label imbalance, as observed in Table \ref{tab:dataset_freq}.

\section{Results}
In this section, we answer the following : 
\begin{itemize}
\setlength\itemsep{1pt}
    \setlength\parsep{0pt}
    \setlength\partopsep{0pt}
    \setlength\leftmargin{10pt}
    \setlength\topsep{1pt}
\item[Q1.] How well does \method{} identify resisting strategies for  Persuasion and Negotiation? 
\item[Q2.] Are resisting strategies good predictors of conversation success? What insights can one glean from the results?
\end{itemize}

\subsection{ Predicting resisting strategies}
We present the results for the automated identification of resisting strategies in Table \ref{Table: Model Results}. We observe that all the models achieve a comparatively lower performance on P4G, mainly due to the higher proportion of `Not-a-Strategy' labels for the latter. 
We gauge the benefits of incorporating conversational context by the significant\footnote{We estimate the statistical significance using the paired bootstrapped test of \citet{berg-kirkpatrick-etal-2012-empirical}, due to the small number of data \cite{dror-etal-2018-hitchhikers}. } improvement of Macro F1 score by 0.036 and 0.011 for P4G and CB respectively. In fact, \method{} outperforms all the proposed baselines significantly.  
%We estimate the statistical significance using the paired bootstrapped test of \citet{berg-kirkpatrick-etal-2012-empirical}, due to the small amount of data \cite{dror-etal-2018-hitchhikers}. 

% The results also highlight the benefits of  incorporating pretrained BERT embeddings, as observed by the steep improvement in macro-F1 score from CNN to BERT-CNN, thereby justifying the use of contextualised embeddings for the task.

\begin{table}[!htp]\centering
\caption{Results of \method{} and other baselines on the resistance strategy prediction task on the Persuasion and CB dataset. The metrics used for evaluation are Macro F1 and Weighted F1 represented as M-F1 and W-F1 respectively. The best results are in bold. }\label{Table: Model Results}
\footnotesize
\resizebox{0.47\textwidth}{!}{
\begin{tabular}{lrrrrr}\toprule
\multirow{2}{*}{Model} &\multicolumn{2}{c}{Persuasion (P4G)} &\multicolumn{2}{c}{Negotiation (CB)} \\\cmidrule{2-5}
&M-F1 &W-F1 &M-F1 &W-F1 \\\midrule
CNN &0.261 &0.757 &0.560 &0.706 \\
BERT + CNN &0.508 &0.819 &0.651 &0.751 \\
HiGRU-sf &0.446 &0.788 &0.605 &0.734 \\
BERT + BiGRU &0.514 &0.815 &0.647 &0.747 \\
BERT + BiGRU-sf &0.522 &0.814 &0.649 &0.750 \\
\textbf{\method{}} &\textbf{0.558} &\textbf{0.828} &\textbf{0.662} &\textbf{0.767} \\
\bottomrule
\end{tabular}}
\end{table}

\begin{figure*}[t]
    \centering
    \includegraphics[scale=0.5]{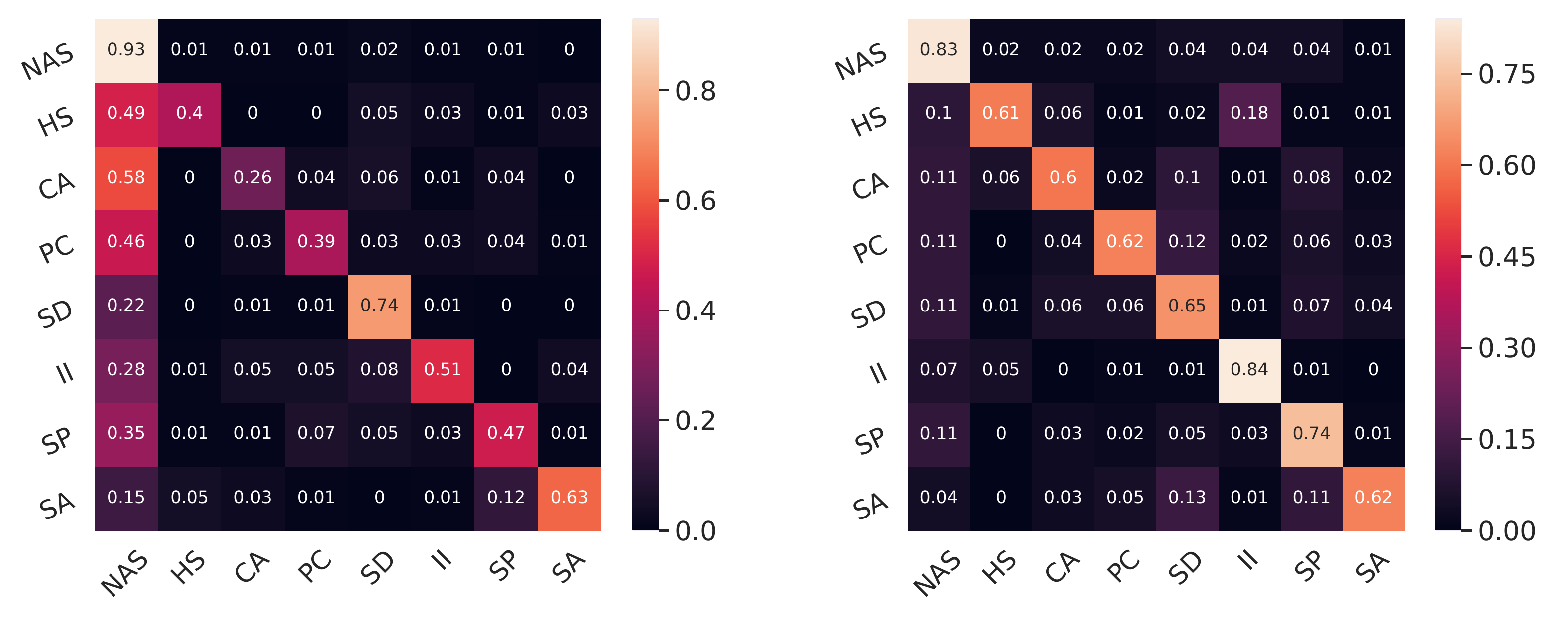}%per_cm.png}
    \caption{Confusion matrix for resisting strategies for the Persuasion (P4G) and Negotiation (CB) datasets on the left and right respectively. Each resisting strategy is represented as its initial (\textbf{S}elf \textbf{P}ity) as SP. True and Predicted Labels have been plotted on the X-axis and the Y-axis respectively.}
    \label{fig:per_neg_cm}
\end{figure*}

% \begin{figure}[t]
%     \centering
%     \includegraphics[scale =0.45]{Figures/resper_negotiation.pdf}%neg_cm.png}
%     \caption{Confusion matrix for resisting strategies for the Negotiation (CB) dataset.}
%     \label{fig:neg_cm}
% \end{figure}

\noindent\textbf{Error Analysis:} We present the confusion matrix for predicting resisting strategies using \method{} on the Persuasion (P4G) and Negotiation (CB) datasets in Figures \ref{fig:per_neg_cm}(a) and \ref{fig:per_neg_cm}(b) respectively. 
We observe that most classification errors occur when a resisting strategy is incorrectly inferred as `Not-A-Strategy'. The effect is more prevalent for P4G since `Not-A-Strategy' comprises 80\% of all annotated labels. Other notable instances of misclassification for P4G occurs when \textit{Self Assertion} is predicted as \textit{Self Pity} since both strategies refer to one's self. These strategies occur so infrequently (see Table \ref{tab:dataset_freq}) that the models lack sufficient information to distinguish between the two categories.
Likewise, for the CB corpus, \textit{Hesitance} utterances which constitute a price request, are often posed as questions. This causes the model to predict the strategy as \textit{Information Inquiry} instead. \textit{Self Assertion} is often incorrectly marked as \textit{Source Derogation} possibly because it often takes a firm stance, and is likely to disparage the other party in the process, thereby confusing the model.

% misclassified as \textit{Information Inquiry},  \textit{Hesitance} often involves a price request, which is usually posed as a question. The example for \textit{Hesitance} in the column for \textit{Negotiation} in Table \ref{tab:framework} would be a good example. An inquiry too often involes an interrogation, thereby confusing the model. 

% \rd{Required to look into this for Nego.}

% \begin{figure}[ht]
%     \includegraphics[scale=0.32]{Figures/per_confusion.jpeg}
%     \caption{Confusion matrix for Persuasion}
%     \label{fig:per-cm.}
%     \includegraphics[scale=0.32]{Figures/nego_confusion.jpeg}
%     \caption{Confusion matrix for Negotiation}
%     \label{fig:neg-cm.}
% \end{figure}

\subsection{Conversation Outcome Prediction}
\label{sec:outcome}

\begin{table}[t]

\caption{We observe the impact of incorporating sequence of strategies on conversation outcome prediction in terms of Macro-F1 and Weighted-F1 score. For P4G, we observe strategies of the persuader (ER), persuadee (EE) and both. For CB, we observe strategies of the buyer (BU), seller (SE) and both.}
\centering
\resizebox{7.5cm}{!}{\begin{tabular}{lcclcc}
\toprule
 \multicolumn{3}{c}{Persuasion (P4G)} & \multicolumn{3}{c}{Negotiation (CB)}  \\

User & Macro-F1 & W-F1 & User & Macro-F1 & W-F1\\
\midrule
ER & 0.588& 0.620 & BU & \textbf{0.618} &\textbf{0.640}\\
EE & 0.618& 0.640 & SE & 0.462& 0.508\\
Both &\textbf{ 0.646}& \textbf{0.671} & Both & 0.605& 0.626\\

% User & Acc & 
% % {}   & Acc   & F1    & Acc   & F1\\

% Strategy & 0.65\pm0.06&   0.62\pm0.05   &0.64\pm0.04&    0.64\pm0.04\\
% Text     & 0.57\pm0.05&    0.55\pm0.05 & 0.64\pm0.04&    0.61\pm0.03 \\
% Strategy + Text &  0.65\pm0.08&    0.64\pm0.06 & 0.64\pm0.05&    0.62\pm0.05 \\
\bottomrule
\end{tabular}}

\label{Table: Outcome Prediction Results}
% \vspace{-0.3cm}
\end{table}

We observe how the sequence of strategies adopted by the two participants have a disproportionate impact on the final conversation outcome in Table \ref{Table: Outcome Prediction Results}. It is interesting to note that the resisting strategies for the persuadee have a greater effect on the conversation outcome (macro-F1 score of 0.62) than the persuasion strategies themselves (macro-F1 score of 0.59). Moreover, incorporating both the persuasion and resisting strategies boosts the prediction performance even further to 0.65. 

We also observe an asymmetry in the roles of the buyer (BU) and the seller (SE) for the CB dataset. We observe that BU's strategies are significantly more effective in deciding the conversation outcome, probably because buyers demonstrate a higher number of resisting strategies.
These experiments highlight the importance of incorporating resisting strategies to gain a complete picture. 

\subsection{Comparative Analysis of Strategies}

Emboldened by the success of resisting strategies to infer the conversational outcome, we probe deeper to investigate the impact of individual strategies. We apply logistic regression with the frequency of strategies, of either participant, as the features while the outcome variable denotes conversation success.  We observe the coefficients of the strategies to infer their correlation with conversation success and their corresponding p-values to determine whether the correlation was indeed statistically significant. Our procedure follows previous work in identifying influential persuasion strategies \cite{diyi-yang-persuasive, persuasion-for-good-2019}. We present the results of this analysis in Table \ref{tab: Coefficients}. 

\begin{table}[!htp]\centering
\caption{Coefficients of the different persuasion strategies corresponding to the persuadee, EE in Persuasion and the buyer, BU, and seller, SE in Negotiation. A value of * and ** means the strategy is signficant with p-value $\leq$ 0.05 and 0.01 respectively. }\label{tab: Coefficients}
\footnotesize
\resizebox{0.47\textwidth}{!}{
\begin{tabular}{lllll}\toprule
&Persuasion (P4G) &\multicolumn{2}{l}{Negotiation (CB)} \\\cmidrule{2-4}
Strategy & EE & BU &SE \\\midrule
Not-A-Strategy &-0.008 &0.287** &-0.138 \\
Hesitance &0.344 &0.328* &0.266 \\
Counter Argument &-0.014 &-0.256 &0.429* \\
Personal Choice &0.153 &0.126 &0.164 \\
Information Inquiry &0.180* &0.091 &-0.704 \\
Source Derogation &0.043 &0.052 &-0.455 \\
Self Pity &0.103 &0.081 &-0.314 \\
Self Assertion &0.843* &-0.576* &-0.040 \\
\bottomrule
\end{tabular}}
\end{table}

For P4G, all the resisting strategies for persuasion apart from \textit{Counter-Argumentation} are positively correlated with a refusal to donate. The highest impact stems from \textit{Self Assertion}. Previous research \cite{fransen2015strategies, zuwerink2003strategies} has noticed that \textit{Self Assertion} is prominent amongst individuals with high self-esteem. Such individuals are confident about their beliefs and less likely to conform. Similarly, a high positive coefficient for \textit{Information Inquiry} can be attributed as follows.  EE inquires information about the charity not only as a means to verify their legitimacy, but also to gain the knowledge they can exploit to their advantage. An innocuous question like `Where will my money go?' would enable EE to assert that they are keener to help children in their own country instead, thereby resisting the donation attempt and saving face. 

The CB scenario setup ensures that the coefficients of the strategies set for BU and SE would be anti-correlated, which holds for the Table \ref{tab: Coefficients}. Like P4G, a high negative coefficient of \textit{Self Assertion} signifies that SE's price is disagreeable to BU - they would instead not buy. Moreover, the high coefficient of \textit{Counter Argumentation} justifies that it is an effective tactic for both parties.

\section{Conclusion}
We present a generalised computational framework grounded in cognitive psychology to operationalise resisting strategies employed to counter persuasion. We identify seven distinct resisting strategies that we instantiate on two publicly available corpora comprising persuasion and negotiation conversations. We adopt a hierarchical sequence labelling architecture to infer the resisting strategies automatically and observe that our model achieves competitive performance for both datasets. Furthermore, we examine the interplay of resisting strategies in determining the final conversation outcome, which corroborates with previous findings. 
In the future, we would like to explore better models to encode the strategy information and apply our framework to improve personalised persuasion and negotiation dialogue systems. We would also like to study the influence of other confounding factors such as power dynamics on the outcomes of conversations featuring resisting strategies.

\section*{Acknowledgments}
We thank the anonymous EACL reviewers for their insightful comments and constructive feedback. This research was funded in part by NSF Grants (IIS 1917668 and IIS 1822831) and Dow Chemical. The first author would also like to acknowledge his best friend, Ahana Sadhu, for her constant support and motivation, who unfortunately and untimely left us this year.

\newpage
\bibliography{references}
\bibliographystyle{acl_natbib}
% \balance
% \newpage
% \beginsupplement
% \balance
\newpage\phantom{blabla}
\newpage\phantom{blabla}
\section*{Appendix}
% The appendix outlines the annotation flowchart for the Negotiation datatset (CB) here. 

\begin{figure}[h]
    \centering
    \includegraphics[width=1.1\textwidth]{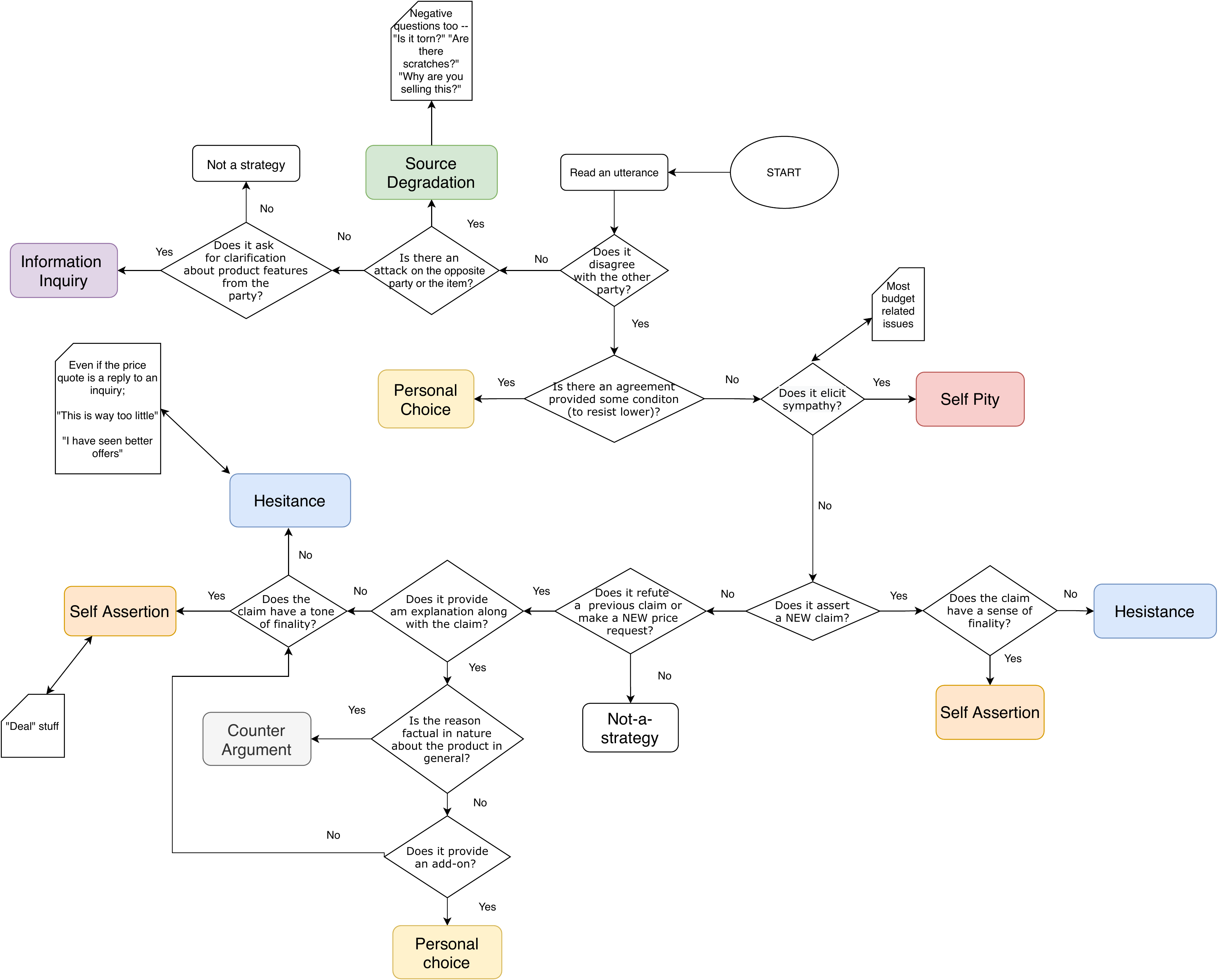}
    \caption{Flowchart for annotating CB}
    \label{fig:flowchart}
\end{figure}

%Please add the following packages if necessary:
%\usepackage{booktabs, multirow} % for borders and merged ranges
%\usepackage{soul}% for underlines
%\usepackage[table]{xcolor} % for cell colors
%\usepackage{changepage,threeparttable} % for wide tables
%If the table is too wide, replace \begin{table}[!htp]...\end{table} with
%\begin{adjustwidth}{-2.5 cm}{-2.5 cm}\centering\begin{threeparttable}[!htb]...\end{threeparttable}\end{adjustwidth}

%Please add the following packages if necessary:
%\usepackage{booktabs, multirow} % for borders and merged ranges
%\usepackage{soul}% for underlines
%\usepackage[table]{xcolor} % for cell colors
%\usepackage{changepage,threeparttable} % for wide tables
%If the table is too wide, replace \begin{table}[!htp]...\end{table} with
%\begin{adjustwidth}{-2.5 cm}{-2.5 cm}\centering\begin{threeparttable}[!htb]...\end{threeparttable}\end{adjustwidth}
% \begin{table}[!htp]\centering
% \caption{Number of parameters for each model in our
% experiments}\label{tab: }
% \scriptsize
% \begin{tabular}{lrr}\toprule
% \textbf{Model} &\textbf{Parameter Size} \\\midrule
% CNN &272708 \\
% BERT + CNN &693908 \\
% HiGRU-sf &14421783 \\
% BERT + BiGRU &13458920 \\
% BERT + BiGRU-sf &16342504 \\
% RESPER &17776407, 15589255 \\
% \bottomrule
% \end{tabular}
% \end{table}

\end{document}